\documentclass[conference]{IEEEtran}
\IEEEoverridecommandlockouts
\usepackage{cite}
\usepackage{amsmath,amssymb,amsfonts}
\usepackage{algorithmic}
\usepackage{graphicx}
\usepackage{float}
\usepackage{xcolor}

\usepackage{multirow}
\usepackage{adjustbox}
\usepackage{url}
\usepackage{hyperref}
\usepackage{booktabs} 
\hypersetup{breaklinks=true}

\usepackage{textcomp}
\usepackage{xcolor}
\def\BibTeX{{\rm B\kern-.05em{\sc i\kern-.025em b}\kern-.08em
    T\kern-.1667em\lower.7ex\hbox{E}\kern-.125emX}}
\begin{document}

\title{To forget is to preserve: Machine Unlearning for 3D medical image segmentation\\
{\footnotesize \textsuperscript{}}
\thanks{}
}

\author{
\IEEEauthorblockN{1\textsuperscript{st} Nitesh Kumar Singh}
\IEEEauthorblockA{\textit{School of Computer Science} \\
\textit{UPES, University of Tomorrow}\\
Dehradun, India. \\
niteshkumar.singh@ddn.upes.ac.in}
\and
\IEEEauthorblockN{2\textsuperscript{nd} Akhilesh Singh}
\IEEEauthorblockA{\textit{School of Computer Science} \\
\textit{UPES, University of Tomorrow}\\
Dehradun, India.\\
akhilesh.19062@stu.upes.ac.in \thanks{\textsuperscript{*}Both 1\textsuperscript{st} and 2\textsuperscript{nd} author contributed equally.}}
\and
\IEEEauthorblockN{3\textsuperscript{rd} Arjun Arora}
\IEEEauthorblockA{\textit{School of Computer Science} \\
\textit{UPES, University of Tomorrow}\\
Dehradun, India. \\
a.arora@ddn.upes.ac.in}

}

\maketitle

\begin{abstract}
\noindent With new data privacy laws such as the General Data Protection Regulation (GDPR) \cite{gdpr2016} that allow individuals to ask that any of their personal information be erased from trained machine learning models, there has been a push to investigate the unlearning of data from models as a way to comply with these laws. 
In this regard, based on four mechanics, we consider several approximate unlearning strategies applied to the MRBrainS18 dataset \cite{mendrik2015mrbrains}. We use a 3D ResNet-50 \cite{he2016deep} as a backbone architecture for segmentation that has been pre-trained with the Med3D framework \cite{chen2019med3d}. 
Considering the pre-trained model as a baseline, we evaluate respective retention accuracy on 2 types of subjects, i.e., retain and forget. We assess these approaches through their Dice similarity coefficient and mean absolute error (MAE) values using two separate training horizons 20 and 50 epochs. The results show that the Noisy Label strategy had the best overall trade-off with a decrease of 93\% in the forget set while maintaining 84\% accuracy for the retained set after 50 epochs.  All other strategies showed extreme levels of forgetting at higher epoch numbers while also demonstrating catastrophic degradation of their retain set performance.  
The results of this study provide a strict baseline of performance metrics for unlearning on a subject-specific level and provide practitioners with clear criteria for selecting the proper strategies.
\end{abstract}


\section{Introduction}
\noindent The domain of medical image analysis has been profoundly altered by deep learning in just the past 10 years. As a result, convolutional neural network (CNN) architectures now provide expert-level performance across a wide variety of clinical tasks from detecting tumors on histopathology slides to delineating organs in computed tomography (CT) volumes \cite{litjens2017survey}. In particular, 3D convolutional architectures have become indispensable for volumetric imaging methods such as magnetic resonance imaging (MRI) and CT because the axial spatial context of 3D volume images contains diagnostically relevant information that two-dimensional, axial slice-by-slice processing cannot fully exploit \cite{ronneberger2015unet}. This realization has driven a growing shift toward fully 3D convolutional architectures that treat the entire volume as a single, coherent input. Works like Med3D \cite{chen2019med3d} have brought this shift into sharp focus, showing that when models are pre-trained across diverse 3D medical datasets, they learn volumetric representations that transfer remarkably well to new clinical tasks, making a compelling case that working natively in 3D is not just a technical nicety, but a genuine step toward building AI systems that can be trusted in real clinical environments.

\medskip

\noindent Despite the remarkable advances in deep learning for medical image analysis, data privacy remains a significant concern. The success of deep neural networks stems in part from their ability to learn and retain detailed information from training data, raising important questions regarding the memorization of patient-specific characteristics \cite{krizhevsky2012imagenet}. Medical imaging datasets inherently contain highly detailed anatomical information at the voxel level \cite{litjens2017survey,ronneberger2015unet}. Variations in brain anatomy, organ morphology, and lesion characteristics may act as quasi-identifiers that can potentially be linked back to individual patients \cite{schwarz2016identifying}. 
To address these concerns, several regulatory frameworks have established requirements for the protection and removal of personal data. In the European Union, the \textit{General Data Protection Regulation (GDPR)} includes the \textit{right to erasure} (Article 17), which requires data controllers to remove personal data upon request and ensure that downstream systems no longer retain information derived from the deleted records \cite{GDPR}. Similar obligations are reflected in the United States through the \textit{Health Insurance Portability and Accountability Act (HIPAA)} \cite{hipaa1996}, and in India through the \textit{Digital Personal Data Protection Act (DPDPA) 2023} \cite{dpdpa2023}.
For healthcare institutions that train deep learning models on patient cohorts, compliance with such regulations extends beyond deleting the original scans from storage. It also requires demonstrating that the trained model no longer retains knowledge attributable to the removed patient's data, thereby motivating the need for effective machine unlearning techniques.

\begin{figure}[H]
    \begin{center}
        \includegraphics[height=5cm, width=8cm]{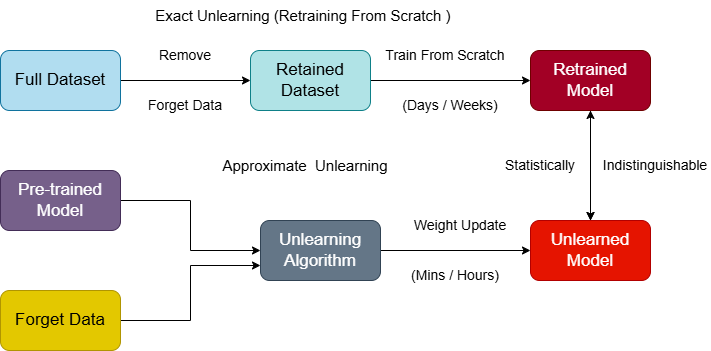}
        \caption{A comparison between exact unlearning (retraining from scratch) and approximate unlearning.}
        \label{fig:ExactApproxUnlearning}
    \end{center}
    \vspace{-0.5cm}
\end{figure}

\medskip
\noindent For achieving this privacy in the data, using a regulatory framework, many research works have been done in the past. One of the methods is a naive approach which retrains the model from scratch on the remaining dataset \cite{cao2015towards, bourtoule2021machine} (see Fig. \ref{fig:ExactApproxUnlearning}). This satisfies the requirement in theory, but is rarely feasible in practice, since deletion requests come in small groups at unknown times, the total cumulative cost of retraining becomes excessive as it can take hundreds of GPU-hours \cite{bourtoule2021machine}. 
Moreover, the cost and computational capacity limitations have driven an increasing amount of work on \textit{machine unlearning} to change an existing model so that it behaves as if the points being deleted were never part of its training data, while avoiding the full cost of the retraining would take. 
The idea of machine unlearning was first introduced in the paper \cite{cao2015towards} as a concept to be solved, framing it as a learning-to-forget issue. The paper \cite{bourtoule2021machine} proposed the use of a Sharded, Isolated, Sliced, and Aggregated (SISA) framework where a deletion request only affects a small portion or shard of the entire training set, thus reducing the total retraining cost. SISA is able to provide verification that a model has truly forgotten data, but it also constrains how the models can be trained originally. Thus, including its principles or limitations are much more difficult if the model has been deployed.

\medskip

\vspace{-0.5cm}

\noindent In recent years, approximate unlearning techniques that directly manipulate the existing pre-trained model by performing targeted weight adjustments to reduce the effects of the forgotten data, have been developed (see Fig. \ref{fig:ExactApproxUnlearning}). The primary goal of these techniques is to create an unlearned model that has a statistical behaviour identical to that of a model which never experienced any training on the forgotten subject, while simultaneously maintaining performance on the retained subject \cite{bourtoule2021machine,golatkar2020eternal}. These approaches allow for more flexibility in both computational power and hardware, typically completed in minutes to hours, but in exchange, they do not have the guarantee of being exact. For instance, the paper \cite{golatkar2020forgetting} proposed weight perturbations to remove class-specific information from neural network parameters based on Fisher information. 
As a result, these alternatives use different hypotheses as to where and how the model stores subject specific knowledge, e.g., encoder features or decoder head, resulting in differing trade-offs between the ability to forget and the retention of utility from the remaining data \cite{graves2021amnesiac}. A detailed discussion about Machine Unlearning is presented below in the Section \ref{Sec:MU}.

\medskip

\noindent While a broad set of tools for evaluating and defining machine unlearning methods exists, all prior evaluation efforts were limited to two-dimensional datasets such as Canadian Institute For Advanced Research-10 (CIFAR-10), ImageNet, Street View House Numbers (SVHN), where a given forget set is comprised of full classes and the evaluation measure with the good accuracy \cite{nguyen2025survey}.
Applying 2D evaluation metrics to the 3D medical imaging data is non-trivial. In clinical imaging data, a given forget set is not a class but a specific subject, i.e., a patient and the output is a dense voxel-wise prediction as opposed to a single class label. Hence the evaluation metrics used in 3D medical imaging datasets are different, for example, in segmentation, it is based on overlap, such as the dice similarity coefficient \cite{milletari2016vnet} and mean absolute error \cite{WillmottMatsuura2005}. Furthermore, the high dimensionality of 3D feature maps substantially increases the risk that aggressive unlearning techniques will lead to the encoder becoming unstable and, subsequently, poor performance across the board \cite{Kirkpatrick2017,golatkar2020eternal}. The various types of difficulties associated with the 3D medical imaging include patient level forgetting sets of data, voxel wise outputs, and volumetrically rich feature spaces. As a result, there has been very little exploration into the unlearning of 3D medical images \cite{Xue2024Erase}. 
Paper \cite{nguyen2025survey} provide a complete overview of the machine unlearning ecosystem but found no benchmarking for dense prediction tasks like segmentation. Liu \cite{liu2022right} investigated federated unlearning in health care using a federated approach where hospitals provide distributed training which complicates removal of data used to train neural networks. Moreover, the paper \cite{becker2021right} examined the right to be forgotten concerning electronic health records and predictive models and indicated the difficulties associated with meeting the requirements outlined in GDPR, particularly Article 17.
Although developing unlearning methods that maintain the structural complexity found in volumetric data represents not only a technical requirement but also a necessary step towards the development of trustworthy clinical AI systems, which respect the principles of privacy and can adequately provide for the demands of providing 3D medical images in a real world healthcare setting.

\medskip

\noindent To address the challenges outlined above, this study builds upon two complementary sources of prior work. First, Med3D \cite{chen2019med3d} provides pre-trained 3D ResNet architectures learned from large-scale multi-organ medical imaging datasets, offering robust volumetric feature representations for downstream medical image analysis tasks. These representations serve as the foundation for investigating the selective removal of patient-specific information from trained segmentation models. Second, the I2I framework proposed by \cite{Li2024Machine} introduces a set of machine unlearning strategies for image-to-image generative models, employing KL-divergence-based objectives to balance knowledge retention and forgetting.
A pre-trained Med3D ResNet-50 backbone was coupled with a segmentation head and fine-tuned on the MRBrainS18 dataset \cite{mendrik2015mrbrains} to establish a baseline segmentation model. Subsequently, several unlearning strategies inspired by \cite{Li2024Machine} were applied to this model, including \textit{Retain Label}/\textit{Max Loss}, \textit{Noisy Label}, \textit{Random Encoder}, \textit{Learn Others}, \textit{Learn Noise}, and \textit{Fix Decoder}. Fig \ref{fig:pipeline} represents the overall pipeline of our work. To investigate this in a controlled setting, we formulate a binary segmentation task on the MRBrainS18 dataset, distinguishing foreground brain tissue from background. These approaches were evaluated and compared across multiple training horizons to assess their ability to balance effective forgetting with preservation of retained knowledge.
\textit{To the best of our knowledge, no prior study has systematically evaluated multiple approximate unlearning strategies for 3D medical image segmentation using subject-level forget sets} \cite{Xue2024Erase}. Our contribution is outlined as follows:
\vspace{-0.6cm}
\begin{figure}[H]
    \begin{center}
        \includegraphics[height=5.5cm, width=8.5cm]{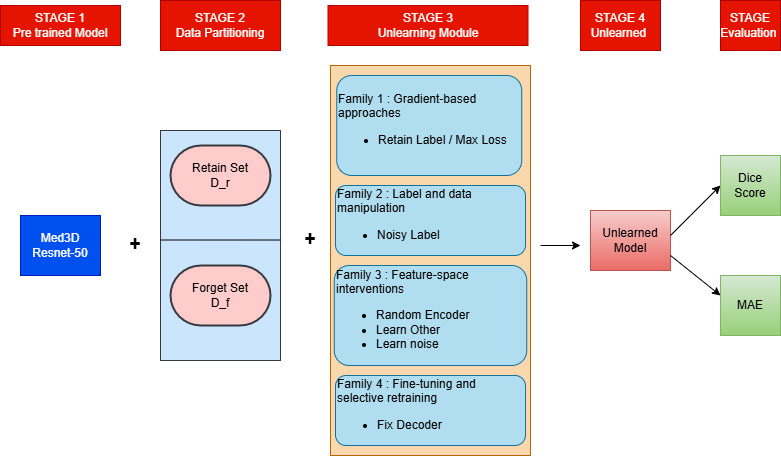}
        \caption{Overall pipeline: Pretrained segmentation model based on Med3D framework together with Retain and Forget data and Unlearning strategies produces Unlearned model.}
        \label{fig:pipeline}
    \end{center}
    \vspace{-0.6cm}
\end{figure}

\noindent \textit{{Contributions:}}
1. We present, to the best of our knowledge, the first systematic benchmark of multiple approximate machine unlearning strategies for {3D medical image segmentation} using {subject-level forget requests}. Unlike existing studies that primarily focus on 2D image classification, our work investigates unlearning in a dense voxel-wise prediction setting representative of real-world medical imaging applications.
\\
2. We adapt several unlearning strategies from the literature to a Med3D ResNet-50 segmentation framework and we present the first comprehensive empirical evaluation of multiple approximate unlearning techniques for removing patient-specific knowledge from 3D medical image segmentation models.
\\
3. We establish a reproducible evaluation protocol for patient-level unlearning in volumetric medical imaging, employing segmentation-specific metrics, including Dice similarity coefficient and mean absolute error, to quantify the trade-off between forgetting effectiveness and retained clinical utility.
\\
4. Through extensive experiments, we characterize the forget-retain behaviour, training-duration effects, and failure modes of different unlearning strategies. Our results identify the \textit{Noisy Label} approach as the most effective method, achieving the strongest forgetting performance while maintaining substantially higher retain-set utility than competing approaches.

\section{RELEATED WORK}

\noindent This section of the paper summarizes important literature, aligning with our work, in two different parts: deep-learning algorithms for volumetric segmentation of medical images, and the foundations of machine unlearning.


\subsection{ Deep Learning for 3D Medical Image Segmentation}
\noindent In the wake of AlexNet's revolutionary success on ImageNet \cite{krizhevsky2012imagenet}, the use of deep convolutional networks in processing medical images became a viable option to a clinical problem very quickly. The paper \cite{ronneberger2015unet} proposed a new type of model called \textit{U-Net}, which is composed of encoders and decoders that are connected via skip connections. U-Net became the accepted method for segmenting biomedical images from a limited amount of data because it provides accurate localization maps from a limited number of training samples. Moreover, the paper \cite{cicek20163dunet} expanded on U-Net by creating a model called \textit{3D U-Net}, which allows volumetric image data such as CT and MRI scans to be processed without losing any of the information present in the original volume.
Apart from this, there is a standard benchmark for brain tissue segmentation using 3D MRI scans of older adults. The MRBrainS18 challenge \cite{mendrik2015mrbrains}, which consists of expert annotations of eight structures (cortical grey matter, basal ganglia, white matter, white matter lesions, cerebrospinal fluid, lateral ventricles, cerebellum, and brainstem), has been frequently used to test classical atlas-based segmentation methods and state-of-the-art deep learning methods. Due to the lack of well-annotated 3D medical datasets, transfer learning has become an interesting technique where researchers used transferred features based on 2D ImageNet to apply to many volumetric data slices by processing each slice individually \cite{shin2016deep}. However, this type of process removes the context that is found between slices.

\medskip

\noindent Paper \cite{chen2019med3d} filled this gap by introducing the Med3D framework which is one of the main facilitator of 3D medical image segmentation research to date. Med3D provides users with a diverse variety of pre-trained 3D ResNet architectures (ResNet-10, 18, 34, 50, 101, 152 and 200) that were jointly trained on eight different datasets of segmentation for lung, liver, and brain anatomy with both CT and MRI modalities. The Med3D pre-trained weight files capture features found in any type of volumetric modality, including volumetric primitives, i.e., edge and texture, and boundary prior features that can be transferred to each subsequent downstream task that will have a limited amount of available labelled datasets for training. 
Med3D can increase the rate of convergence compared to Kinetics-based pre-training by 2 times and 10 times as compared to training from scratch, while improving accuracy for segmentation of the target datasets from 3–20 percentage points, depending upon the target task. 


\subsection{Foundations of Machine Unlearning}
\label{Sec:MU}

\noindent Machine unlearning was first formally studied by Cao and Yang \cite{cao2015towards}, who described the problem as reformulating an existing model so that the output of that model is statistically equivalent to the output of a model that was not ever trained with the relevant data. They cast the learning algorithm into a summation form so that it would allow for efficient computation of how to subtract the contribution of an individual point of data from the model, but their methodology was only applicable to very narrowly defined families of models, for example, naive Bayes or other linear classifiers. 
Moreover, the paper \cite{bourtoule2021machine} introduced a methodology called SISA to accomplish exact unlearning in an arbitrary model through restructuring the training process. The SISA framework partitions the training set into disjoint shards and each shard is independently trained using a different constituent model. The final results at inference are based on aggregating the results of all of the constituent models. Therefore, if a deletion request is made for an empirical data point, only the constituent model for that shard needs to be retrained; however, intra-shard slicing with checkpointing can further limit the number of models needing retraining. There are many theoretical guarantees associated with the SISA framework, however, there is also an associated cost to using this methodology: The training infrastructure must be configured for sharding from the beginning and retrofitting SISA to an existing model is not practical. Additionally, when aggregating the results of many separate (but dependent) constituent models, there is a significant increase in latency and memory usage, both of which can be problematic in a resource-constrained clinical inference setting.
Furthermore, the paper \cite{ginart2019making} studied how to unlearn precisely using k-means clustering and showed that efficient data removal can be done via the combination of a learning algorithm with a specified algebraic structure. However, deep neural networks used to segment medical images do not have that algebraic structure, making it impossible to unlearn precisely mathematically from those methods.

\medskip

\noindent As a result of the limitation of exact unlearning methods, a significant number of studies have examined approximate strategies for unlearning and have been able to achieve some degree of computational feasibility in terms of the theoretical guarantees offered by these new approximate unlearning strategies. As described in \cite{Li2024Machine}, approximate unlearning has been adapted to the area of generative models 
where the authors set out to minimize the KL-divergence between a forget dataset compared with a set of retained datasets. Their method produces high-quality reconstruction from large datasets like ImageNet-1K and Places-365, but do not require access to the original source images during the training process for the corresponding generative model. These techniques are important because they offer practical unlearning techniques that can also be applied in general, non-generative model-based methods, including medical image segmentation in 3D, as discussed further in this paper. We go over different types of approximate unlearning methods, detailed in Section \ref{Sec:III.C} and discuss how they each work to allow selective removal of known data from their effects on the performance of their final retained machine learning model.

\section{PROPOSED METHODOLOGY}
\noindent This section provides an overview of the general structure of subject-specific machine unlearning applied to 3D medical imaging segmentation. Starting from defining the problem statement, followed by the backbone architecture, then the unlearning strategies that we utilized, and finally the evaluation method selected.

\subsection{Problem statement}

\noindent Let \(\mathcal{D} = \{(x_i, y_i, s_i)\}_{i=1}^{N}\) denote a training set of \(N\) 3D data, where \(x_i \in \mathbb{R}^{1 \times D \times H \times W}\) represents the intensity volume, \(y_i \in \{0, 1\}^{C \times D \times H \times W}\) denotes the one-hot encoded voxel-wise ground-truth segmentation for \(C\) classes, and \(s_i \in \{-1, 1\}\) is the subject identifier. A segmentation model \(f_{\theta} : \mathbb{R}^{1 \times D \times H \times W} \rightarrow \mathbb{R}^{C \times D \times H \times W}\), parameterized by weights $\theta$, outputs raw logits \(z_i = f_{\theta}(x_i)\). The model is trained on \(\mathcal{D}\) to minimize the voxel-wise cross-entropy loss \cite{entropy27040368}:
\vspace{-0.2cm}
\begin{equation*}
\mathcal{L}_{\mathrm{CE}}(\theta) = -\frac{1}{N} \sum_{i=1}^{N} \frac{1}{|\Omega|} \sum_{v \in \Omega} \sum_{c=0}^{C-1} y_{i, c}^{v} \log \big( \sigma(f_{\theta}(x_i))_{c}^{v} \big),
\end{equation*}
where \(v \in \Omega\) indexes the voxels in the spatial domain (\(D \times H \times W\)), \(|\Omega|\) is the total number of voxels per volume (optional, used for normalization), and \(\sigma(\cdot)_c^v\) denotes the softmax probability calculated for class \(c\) at voxel \(v\):
\vspace{-0.2cm}
\begin{equation*}
\sigma(f_{\theta}(x_i))_{c}^{v} = \frac{e^{f_{\theta}(x_i)_{c}^{v}}}{\sum_{j=0}^{C-1} e^{f_{\theta}(x_i)_{j}^{v}}}.
\end{equation*}

\noindent Given a forget set $\mathcal{D}_f \subset \mathcal{D}$ which consists of all the  samples which belongs to designated forget subjects, and a retain set $\mathcal{D}_r = \mathcal{D} \setminus \mathcal{D}_f$, where the goal of a machine unlearning is to obtain the modified model with weights $\theta^*$ such that:\\
\noindent 1. \textit{Forget Criteria:} $f_{\theta^*}$ performs poorly on $\mathcal{D}_f$, which idicates that the complete removal of learned information.\\
\noindent 2. \textit{Retain Criteria:} $f_{\theta^*}$ maintains the high performance on $\mathcal{D}_r$, which ideally matched the original $f_{\theta}$ and the model that retrain on $\mathcal{D}_r$.
\subsection{Backbone architecture}
\vspace{-0.2cm}
 \begin{figure}[H]
    \begin{center}
        \includegraphics[height=6cm, width=8.5cm]{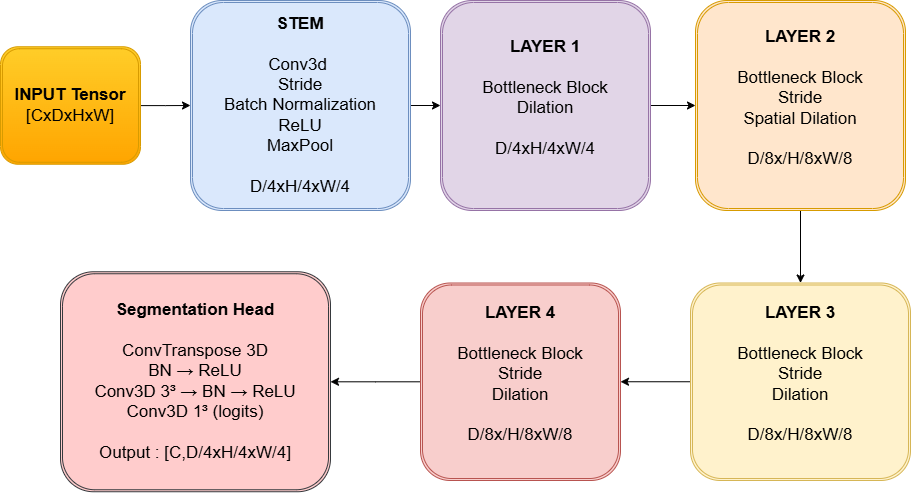}
    \caption{Backbone architecture: Med3D ResNet 50 + Segmentation head}
    \label{fig:Med3DResnet}
    \end{center}
    \vspace{-0.5cm}
\end{figure}
\noindent We explicitly decompose the segmentation model $f_\theta$ into two cascading components: an encoder network $E_\phi$ that extracts latent spatiotemporal features, and a decoder/segmentation head $G_\psi$ that transforms these features into voxel-wise class logits, such that $f_\theta(x_i) = G_\psi(E_\phi(x_i))$ with parameters $\theta = \{\phi, \psi\}$. As segmentation backbone we are utilizing the Med3D ResNet-50 \cite{chen2019med3d, he2016deep}, which extends the ResNet architecture to 3D (see Fig. \ref{fig:Med3DResnet}). The network consists of an input layer followed by 3D convolutional stem (a single  $7 \times 7 \times 7$ convolutional layer, where stride is 2  Batch Normalization Layer, ReLU, and Max-Pooling Layers), then after by four residual stages described in layers $(\mathrm{Layer}\,1\text{--}\mathrm{Layer}\,4)$ having 256, 512, 1024, and 2048 channels, respectively. Each of the four residual stages will utilize a group of stacked paired bottleneck blocks with identity short-cut connections of type B (projection to the dimension of the paired residuals' channels). In the final stage of our backbone {we} have a segmentation head which maps the 2048-channel feature maps outputted from Layer 4 to C channels of output via a multi-layer decoder comprising a 3D transposed convolution (2048→32, kernel=2, stride=2) for spatial upsampling, followed by batch normalization, ReLU, a $3 \times 3 \times 3$ refinement convolution $(32\rightarrow32)$, batch normalization, ReLU, and a final 1×1×1 convolution (32→C) for per-voxel class prediction.

\vspace{-0.1cm}
\subsection{Unlearning strategies}
\label{Sec:III.C}

\noindent In this section, based on four mechanics (see Fig. \ref{fig:Unlearningfamilies}), we discuss several different unlearning strategies, adapted from \cite{Li2024Machine}. They are detailed as follows:
\vspace{-0.2cm}
\begin{figure}[H]
    \begin{center}
        \includegraphics[height=7.5cm, width=8.8cm]{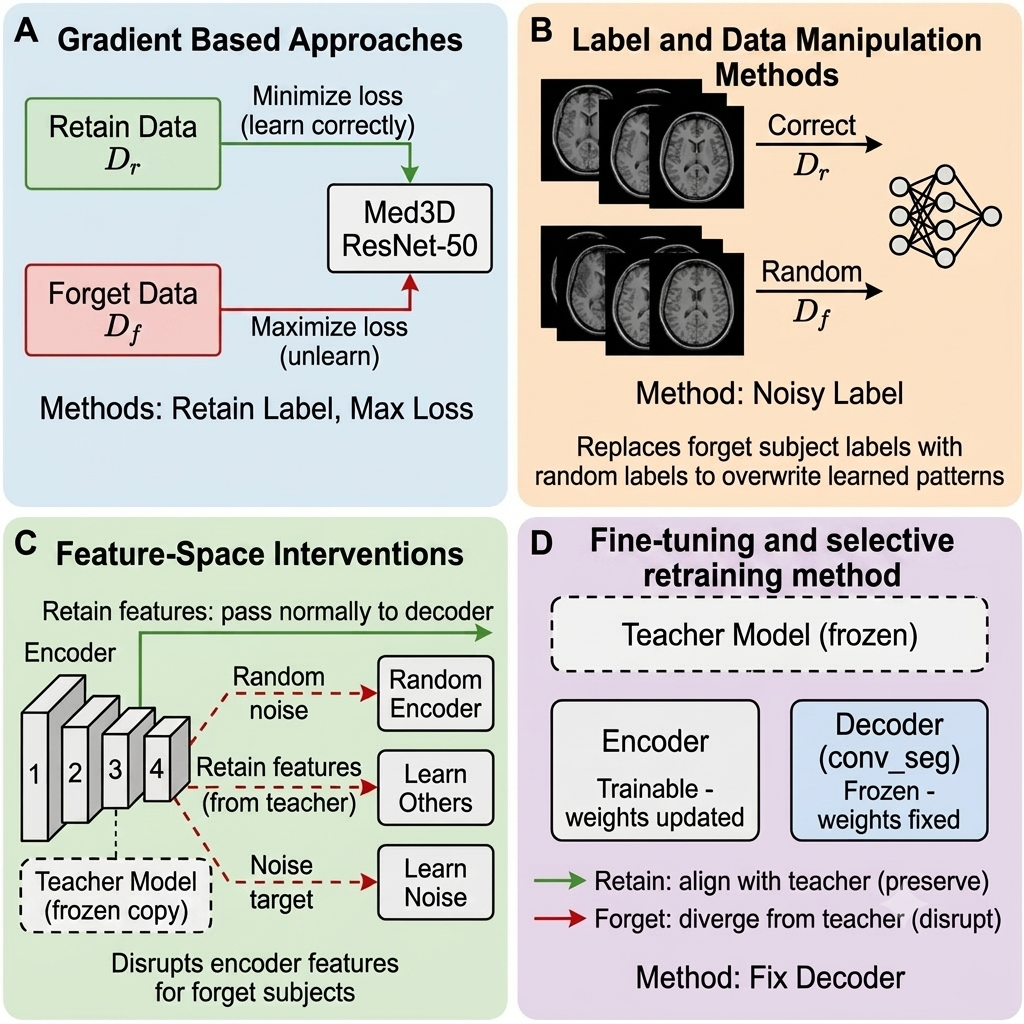}
        \caption{ Approximate unlearning strategies}
        \label{fig:Unlearningfamilies}
    \end{center}
    \vspace{-0.4cm}
\end{figure}
\subsubsection{Gradient-based approaches (GA)}
\noindent This strategy combines the classical gradient-reversal approach, often referred to as Retain Label (RL) or Negative Gradient Learning \cite{bourtoule2021machine, Li2024Machine}, with the adversarial Max Loss (ML) formulation \cite{warnecke2023machine}. However, unbounded neg-grad-ascent can lead to degradation in performance because it can end up driving the model into bad regions of parameter space that cause performance degradation for both forget and retain data \cite{jia2023model}. It creates a clean optimization separation between the retain and forget objectives by modifying the sign of the cross-entropy loss for the forget subjects while preserving standard minimization for the retain subjects:
\vspace{-0.1cm}
\begin{equation*}
\mathcal{L}_{\mathrm{GA}}(\theta) = \mathcal{L}_{\mathrm{CE}}(\theta; \mathcal{D}_r) - \alpha \, \mathcal{L}_{\mathrm{CE}}(\theta; \mathcal{D}_f),
\end{equation*}
where $\alpha > 0$ is a hyperparameter scaling the intensity of the forget gradient, and $\mathcal{L}_{\mathrm{CE}}(\theta; \mathcal{D}_{\text{sub}})$ represents established voxel-wise cross-entropy loss evaluated explicitly over a specific dataset subset $\mathcal{D}_{\text{sub}}$:
\begin{align*}
& \mathcal{L}_{\mathrm{CE}}(\theta; \mathcal{D}_{\text{sub}}) \\
& = -\frac{1}{|\mathcal{D}_{\text{sub}}| \cdot |\Omega|} \sum_{i: (x_i, y_i, s_i) \in \mathcal{D}_{\text{sub}}} \sum_{v \in \Omega} \sum_{c=0}^{C-1} y_{i, c}^{v} \log \big( \sigma(f_{\theta}(x_i))_{c}^{v} \big).
\end{align*}

\noindent By implicitly normalizing each component by its respective subset cardinality ($|\mathcal{D}_r|$ and $|\mathcal{D}_f|$) inside $\mathcal{L}_{\mathrm{CE}}$, this formulation prevents the larger sample size of the retain set from overshadowing the unlearning signal. Negation of the forget loss forces the optimizer to maximize the prediction error on the designated forget subjects, thereby effectively reversing the features learned from those samples.

\subsubsection{Label and data manipulation methods}
In contrast to methods that alter the optimization objective, the Noisy Label (NL) strategy changes the underlying target distribution by substituting the voxel-wise ground-truth class labels of the forget subjects with uniformly random labels \cite{chen2022recommendation, Li2024Machine}. We define the manipulated target tensor $\tilde{y}_i \in \{0, 1\}^{C \times D \times H \times W}$ for sample $i$ as follows:
\begin{equation*}
\tilde{y}_{i, c}^{v} \!=\!
\begin{cases}
y_{i, c}^{v}, & \!\!\text{if } s_i = 1 \ (\text{i.e., } (x_i, y_i, s_i) \in \mathcal{D}_r), \\
\text{OneHot}(\mathbf{c}_r)_{c}^{v}, &\!\! \text{if } s_i = -1 \ (\text{i.e., } (x_i, y_i, s_i) \in \mathcal{D}_f),
\end{cases}
\end{equation*}
where $\mathbf{c}_r \sim \text{Uniform}(\{0, 1, \dots, C-1\})$ is a random class index drawn uniformly at each voxel position $v \in \Omega$ independently across training epochs, and $\text{OneHot}(\cdot)$ transforms this scalar index into a $C$-dimensional binary indicator vector.

\medskip

\noindent The unlearning optimization process minimizes the standard voxel-wise cross-entropy loss over the modified dataset distribution $\tilde{\mathcal{D}} = \{(x_i, \tilde{y}_i, s_i)\}_{i=1}^{N}$:
\begin{equation*}
\mathcal{L}_{\mathrm{NL}}(\theta) = \mathcal{L}_{\mathrm{CE}}(\theta; \tilde{\mathcal{D}}_r) + \mathcal{L}_{\mathrm{CE}}(\theta; \tilde{\mathcal{D}}_f),
\end{equation*}
where $\tilde{\mathcal{D}}_r$ and $\tilde{\mathcal{D}}_f$ represent the partitioned subsets containing the original retain labels and the newly manipulated forget noisy labels, respectively. By executing backpropagation under this formulation, the network incrementally updates and overwrites its memorized representations for the forget subjects due to the fluctuating random label assignments presented at every epoch. Simultaneously, the model continues to receive the correct label coordinates for the retain subjects, effectively reinforcing existing learned structures while selectively inducing catastrophic forgetting on the designated forget subjects.



\subsubsection{Feature-Space Interventions}

\noindent Rather than modifying the optimization loss signs or altering the output ground-truth labels, an alternative paradigm manipulates the internal hidden representations or latent feature maps within the model's architecture. Recent frameworks, such as SalUn \cite{fan2024salun}, utilize gradient-based saliency mapping to isolate and perturb specific latent weights. Following the architectural intervention principles outlined in \cite{Li2024Machine}, we explore three distinct representation-masking and distillation strategies.\\
(i) \textit{Random Encoder (RE)}:
The Random Encoder strategy forces the network to decouple its latent spaces from the anatomy of the forget-set by overwriting the hidden features of forget subjects with Gaussian noise prior to decoding \cite{Li2024Machine}. Specifically, the input feature map $z_i$ passed to the decoder $G_\psi$ is formulated as:
\begin{equation*}
z_i =
\begin{cases}
E_\phi(x_i), & \text{if } s_i = 1 \ (\text{i.e., } (x_i, y_i, s_i) \in \mathcal{D}_r), \\
\mathbf{\epsilon}_i \sim \mathcal{N}(\mu_i, \sigma_i^2), & \text{if } s_i = -1 \ (\text{i.e., } (x_i, y_i, s_i) \in \mathcal{D}_f),
\end{cases}
\end{equation*}
where $\mu_i$ and $\sigma_i$ represent the channel-wise spatial mean and standard deviation computed directly from the true latent feature volume $E_\phi(x_i)$ to preserve structural feature statistics. The model is subsequently optimized by minimizing the standard voxel-wise cross-entropy loss over the manipulated latent stream:
\vspace{-0.1cm}
\begin{align*}
& \mathcal{L}_{\mathrm{RE}}(\theta) = \mathcal{L}_{\mathrm{CE}}(\theta; \mathcal{D}_r) \\
& \quad - \frac{1}{|\mathcal{D}_f| \cdot |\Omega|} \sum_{i: s_i=-1} \sum_{v \in \Omega} \sum_{c=0}^{C-1} y_{i, c}^{v} \log \big( \sigma(G_\psi(z_i))_{c}^{v} \big).
\end{align*}
By backpropagating through this noisy representation, the decoder is forced to process uninformative spatial variations for forget subjects, effectively erasing subject-specific feature dependencies.\\
(ii)  \textit{Learn Others (LO) and Learn Noise (LN)}:
The Learn Others (LO) and Learn Noise (LN) strategies both leverage a teacher-student distillation paradigm to eliminate targeted anatomical characteristics from the latent space \cite{Li2024Machine}. A pre-trained, frozen copy of the network serves as the reference teacher model, denoted as $f_{\theta_T} = G_{\psi_T} \circ E_{\phi_T}$. For all valid retain subjects, the student encoder $E_\phi$ is optimized to directly mirror the teacher's original feature extractions. However, for forget subjects, the student encoder is forced to map inputs to an uninformative, generic latent distribution. 
The two strategies are differentiated strictly by their definition of the forget target within the distillation target tensor $t_i$:
\vspace{-0.2cm}
\begin{equation*}
t_i =
\begin{cases}
E_{\phi_T}(x_i), & \text{if } s_i = 1 \ (\text{i.e., } (x_i, y_i, s_i) \in \mathcal{D}_r), \\
\mathcal{T}_{\mathrm{forget}}(x_i), & \text{if } s_i = -1 \ (\text{i.e., } (x_i, y_i, s_i) \in \mathcal{D}_f),
\end{cases}
\vspace{-0.2cm}
\end{equation*}
where the forget mapping function $\mathcal{T}_{\mathrm{forget}}(x_i)$ is specific to each strategy:\\
a.  LO sets the target to a static anatomical average across the active mini-batch of remaining retain subjects, wiping out patient-specific biometric structures:
\vspace{-0.2cm}
    \begin{equation*}
    \mathcal{T}_{\mathrm{forget}}^{\mathrm{LO}}(x_i) = \frac{1}{|\mathcal{D}_r^{\text{batch}}|} \sum_{j \in \mathcal{D}_r^{\text{batch}}} E_{\phi_T}(x_j).
    \vspace{-0.2cm}
    \end{equation*}
b. LN sets the target to a dynamic, uninformative Gaussian noise field parameterized by the running statistics of the active student encoder:
\vspace{-0.2cm}
    \begin{equation*}
    \mathcal{T}_{\mathrm{forget}}^{\mathrm{LN}}(x_i) \sim \mathcal{N}\left(\mu_{E_{\phi}(x_i)}, \sigma_{E_{\phi}(x_i)}^{2}\right).
    \vspace{-0.2cm}
    \end{equation*}
To align the student parameters with these targets, the student encoder minimizes a Mean Squared Error (MSE) feature alignment objective:
\vspace{-0.2cm}
\begin{align*}
\mathcal{L}_{\mathrm{LO / LN}}(\phi) =& \frac{1}{|\mathcal{D}_r|} \sum_{i: s_i=1} \left\lVert E_{\phi}(x_i) - t_i \right\rVert_2^2 \\
& + \alpha \frac{1}{|\mathcal{D}_f|} \sum_{i: s_i=-1} \left\lVert E_{\phi}(x_i) - t_i \right\rVert_2^2,
\vspace{-0.2cm}
\end{align*}
where $\alpha > 0$ balances the unlearning velocity relative to utility retention. Minimizing this objective successfully collapses the unique structural signatures of forget-set profiles into generic or random feature matrices, while sustaining rigid, standard alignments for the remaining retain distribution.

\subsubsection{Fine-tuning and selective retraining method}
An alternative class of unlearning strategies focuses on updating a constrained subset of the model parameters or employing target-driven distillation to isolate the unlearning effects. For instance, fast unlearning techniques combine capable and incompetent teacher models to induce near-zero accuracy on the forget set while maintaining performance on the retain set at a fraction of the computational cost of full retraining \cite{tarun2023fast}. Similarly, amnesiac unlearning tracks and reverses specific parameter updates linked to the forget data \cite{graves2021amnesiac}. While conceptually elegant, caching every intermediate update is computationally impractical for long training runs on 3D volumetric medical data. Instead, selective fine-tuning can be achieved by locking specific architectural layers to isolate parametric updates.
The Fix Decoder (FD) strategy constraints the optimization trajectory by freezing the segmentation head while updating only the encoder parameters via teacher-student distillation \cite{hinton2015distilling, Li2024Machine}. By setting the gradients of the decoder to fixed values ($\text{requires\_grad} = \text{False}$), the parameters $\psi$ are locked, and a pre-trained, frozen copy of the complete model serves as the teacher reference $f_{\theta_T}(x_i)$. The student encoder parameters $\phi$ are optimized to minimize the following:
\vspace{-0.16cm}
\begin{align*}
\mathcal{L}_{\mathrm{FD}}(\phi) = &  \frac{1}{|\mathcal{D}_r|} \sum_{i: s_i=1} \left\lVert f_{\theta}(x_i) - f_{\theta_T}(x_i) \right\rVert_2^2 \\
& - \alpha \frac{1}{|\mathcal{D}_f|} \sum_{i: s_i=-1} \left\lVert f_{\theta}(x_i) - f_{\theta_T}(x_i) \right\rVert_2^2,
\vspace{-0.3cm}
\end{align*}
where $\alpha > 0$ balances the unlearning rate, and $\lVert \cdot \rVert_2^2$ computes the mean squared error over the output logits across the voxel spatial domain $\Omega$. Freezing the decoder confines the forgetting entirely to the latent feature representations produced by the encoder $E_\phi$. This architectural constraint prevents catastrophic failure stemming from decoder drift or total representation collapse, which often occurs in unconstrained unlearning methods. 

\subsection{Evaluation Criteria}
\noindent Upon completion of the unlearning phase, the updated model parameters $\theta^*$ are evaluated across all available patient subjects. To quantify unlearning performance and utility retention, the evaluation metrics are partitioned into two distinct domains: the retain dataset subset ($\mathcal{D}_r$) and the forget dataset subset ($\mathcal{D}_f$). 

\subsubsection{Dice Similarity Coefficient (DSC)}
The Dice Similarity Coefficient (DSC) is employed to calculate the spatial overlap and semantic similarity between the predicted segmentation mask and the ground-truth annotations. For a specific class $c \in \{0, 1, \dots, C-1\}$, the voxel-wise DSC for sample $i$ is:
\vspace{-0.2cm}
\begin{equation*}
\mathrm{Dice}_c = \frac{2 \displaystyle\sum_{v \in \Omega} \hat{y}_{i, c}^v \cdot y_{i, c}^v}{\displaystyle\sum_{v \in \Omega} \hat{y}_{i, c}^v + \displaystyle\sum_{v \in \Omega} y_{i, c}^v},
\vspace{-0.2cm}
\end{equation*}
where $y_{i, c}^v \in \{0, 1\}$ is the one-hot encoded ground-truth indicator for class $c$ at voxel $v$, and $\hat{y}_{i, c}^v = \mathbf{1}\big(c = \arg\max_j \sigma(f_{\theta^*}(x_i))_j^v\big)$ represents the argmax-predicted binary class assignment. In this study, we evaluate binary segmentation ($C = 2$), tracking the mean DSC exclusively for the foreground target class ($c = 1$). A DSC value of $1$ indicates perfect spatial correspondence, whereas $0$ represents absolute non-overlap. A successful unlearning trajectory must maintain a high, stable $\mathrm{Dice}(\mathcal{D}_r)$ to preserve model utility, while driving $\mathrm{Dice}(\mathcal{D}_f)$ toward $0$ to signify the removal of the forgotten subjects' features.

\subsubsection{Mean Absolute Error (MAE)}
To measure the subtle distribution changes in continuous prediction probabilities rather than hard thresholded assignments, the MAE computes the voxel-level classification error across the probability space:
\vspace{-0.2cm}
\begin{equation*}
\mathrm{MAE} = \frac{1}{C \cdot |\Omega|} \sum_{c=0}^{C-1} \sum_{v \in \Omega} \left| \sigma(f_{\theta^*}(x_i))_c^{v} - y_{i, c}^v \right|.
\vspace{-0.2cm}
\end{equation*}
The MAE quantifies the average per-voxel, per-class absolute deviation between the predicted softmax probability distribution and the true one-hot encoded target configuration. An elevated MAE computed over the forget subset ($\mathcal{D}_f$) implies that the network's confidence in accurately representing the forgotten anatomy has been thoroughly degraded.

\subsubsection{Unlearning Efficacy Metrics}
To assess the global quality and specificity of the unlearning process, we define two composite gap metrics that contrast performance between the retained and forgotten cohorts:
\\
\textbf{Dice Gap:} $\Delta_{\mathrm{Dice}} = \mathrm{Dice}(\mathcal{D}_r) - \mathrm{Dice}(\mathcal{D}_f)$. A large positive gap indicates highly effective selective forgetting. The baseline model prior to unlearning exhibits a gap of $-0.017$, indicating that the forget subjects originally outperformed the retain subjects slightly.
\\
\textbf{MAE Gap:} $\Delta_{\mathrm{MAE}} = \mathrm{MAE}(\mathcal{D}_f) - \mathrm{MAE}(\mathcal{D}_r)$. A positive gap demonstrates that the model actively expresses higher prediction error and lower confidence exclusively on the forget subjects. The pre-unlearning baseline gap is $-0.010$.

\medskip

\noindent An optimal unlearning algorithm maximizes both $\Delta_{\mathrm{Dice}}$ and $\Delta_{\mathrm{MAE}}$ concurrently.

\section{EXPERIMENTS AND RESULTS}

\noindent Our all experiments are performed on NVIDIA H100 80GB HBM3 GPUs. The implementation uses PyTorch with CUDA for acceleration and multi-GPU support through DataParallel.

\subsection{Data setup and implementation details} 

\noindent \textbf{Dataset.} {We have utilized the MRBrainS18 dataset \cite{mendrik2015mrbrains} that has completely independent data from 10 different 3T MRIs obtained from 10 different elderly subjects, all with professional-quality segmentation. For unlearning experiments, two of the subjects only (74.nii.gz and 75.nii.gz) were selected as the unlearned subjects $\mathcal{D}_f$, while each of the remaining sets was assigned as a learned subject $\mathcal{D}_r$. This 2/10 (20\% unlearned ratio) scenario of unlearning clinical data is an approximate representation of an actual situation in which a few of the patients have withdrawn their consent to participate in research.}
{Every volume has been loaded and normalized as a float32 tensor, where the means have all been centered around 0 and all standard deviations are equal to 1. The ground truth labels will be extracted from the volume in binary format to allow for segmentation into two classes, which will be used for $C = 2$ segmentation of the dataset (with 1 class being the brain and other class being non-brain tissue). The dataset loader uses from each sample a forget flag ($s_i = -1$ to forget and $s_i = 1$ to keep the subject) that determines how the unlearning strategies use the different treatment on the different samples in the dataset.}

\medskip

\noindent \textbf{Fine-tune.} {To achieve the Med3D pretraining resolution, we resized the input volume to $56 \times 448 \times 448$ (D x H x W) averaging about 43 MB per sample in float32. Despite there being a small batch size, the stochastic mixing of retain and forget subjects throughout each epoch produces sufficient gradient diversity to achieve stable optimization using most strategies. Next, we initialized the encoder using the Med3D pre-trained weights that were co-trained on the 3DSeg-8 multi-domain dataset \cite{chen2019med3d} and then we {fine-tuned the entire network (encoder + segmentation head) on MRBrainS18 \cite{mendrik2015mrbrains} for 110 epochs using SGD with momentum $0.9$ at a learning rate and weight decay of $1 \times 10^{-3}$.} Hence the resulting model achieved a mean dice coefficient of 0.9139 and MAE of 0.0416 amongst all ten training subjects and this was used as our baseline for future unlearning experiments (see Table \ref{tab:unlearning}).}

\begin{table}[t]
\centering
\caption{Unlearning hyperparameter configuration}
\label{Tab:config}
\begin{tabular}{|l|l|}
\hline
\textbf{Parameter} & \textbf{Value} \\ \hline
Training horizons & 20 and 50 epochs \\ \hline
Optimizer & Adam \\ \hline
Learning rate & $1 \times 10^{-4}$ \\ \hline
Weight decay & $1 \times 10^{-4}$ \\ \hline
LR scheduler & Exponential ($\gamma = 0.99$) \\ \hline
Batch size & 1 \\ \hline
Input dimensions & $56 \times 448 \times 448$ \\ \hline
Number of classes & 2  \\ \hline
$D_f$ & $ 2/10 \; $ \\ \hline
$\alpha$  & 0.5 \\ \hline
\end{tabular}
\vspace{-0.5cm}
\end{table}

\noindent \textbf{Training.} To begin the unlearning phase, we took entire fine-tuned network and then apply the unlearning strategies directly to the neural network. To control for their impact on the gradients and to avoid catastrophic forgetting of the subjects that remain in the database, we switched to using an Adam optimizer, and reduced our learning rate and weight decay to $1 \times 10^{-4}$. We performed the unlearning procedure for either 20 or 50 epochs using a constant learning rate across all layers. For strategies that required loss balancing, we used an alpha ($\alpha$) hyperparameter that was typically set to 0.5 to apply weights to the unlearning objective versus the utility preservation of the retained subjects.
Our hyperparameter settings are summarized in the Table \ref{Tab:config}.


\subsection{Result discussion}

\noindent Based on four mechanics, Table \ref{tab:unlearning} shows that several unlearning strategies exhibit markedly different trade-offs between retain-set preservation and forget-set removal. This table shows the training time in sec., the dice score and MAE for $D_r$ and $D_f$, dice gap and MAE gap for two different training horizons, i.e., $20$ and $50$. We conclude that a given strategy is good if it gets the largest dice gap and MAE gap. Our results are:
\begin{itemize}
    \item Among all methods, \textit{Noisy Label} achieved the strongest selective unlearning performance, producing the largest Dice gap (+0.6963) and MAE gap (+0.091) at 50 epochs while maintaining a retain Dice of 0.7576, reducing the forget Dice to 0.0613, decreasing the retain MAE to 0.0851 and increasing the forget MAE to 0.1761. This method decreases the dice for the forget set by 93\% (0.9226 to 0.0613), while retaining the most percentage of accuracy for the retain set, i.e., 84\%. Moreover, unlike the other approaches, its performance continued to improve with longer training, simultaneously strengthening retention and forgetting.
    \item \textit{Retain Label} or \textit{Max Loss}, achieves strong retain performance after 20 epochs (approximately 0.86 retain Dice and 0.066 retain MAE) but suffered severe over-unlearning for the forget set. The situation is drastically bad when trained for 50 epochs although both methods completely removed forget-set performance, they also caused substantial degradation of retain-set accuracy, demonstrating the instability of unconstrained loss-negation approaches.
    \item \textit{Random Encoder} preserved most retain performance but achieved only limited forgetting with a dice gap of +0.04 (20 epochs), +0.01 (50 epochs) and MAE gap +0.0065 (20 epochs), +0.0026 (50 epochs). This indicates that the decoder could compensate for corrupted encoder features. In contrast, \textit{Learn Others} and \textit{Learn Noise} caused near-complete collapse of retain performance, suggesting that aggressive feature-space manipulation at the chosen learning rate severely disrupted learned representations.
    \item The \textit{Fix Decoder} approach demonstrated moderate unlearning at 20 epochs but became unstable with extended training, eventually reversing the intended behaviour by producing higher performance on forget samples than on retain samples. This failure mode highlights the risks associated with imposing architectural constraints during unlearning.
\end{itemize}

\begin{figure}[H]
    \begin{center}
        \includegraphics[height=5cm, width=8.8cm]{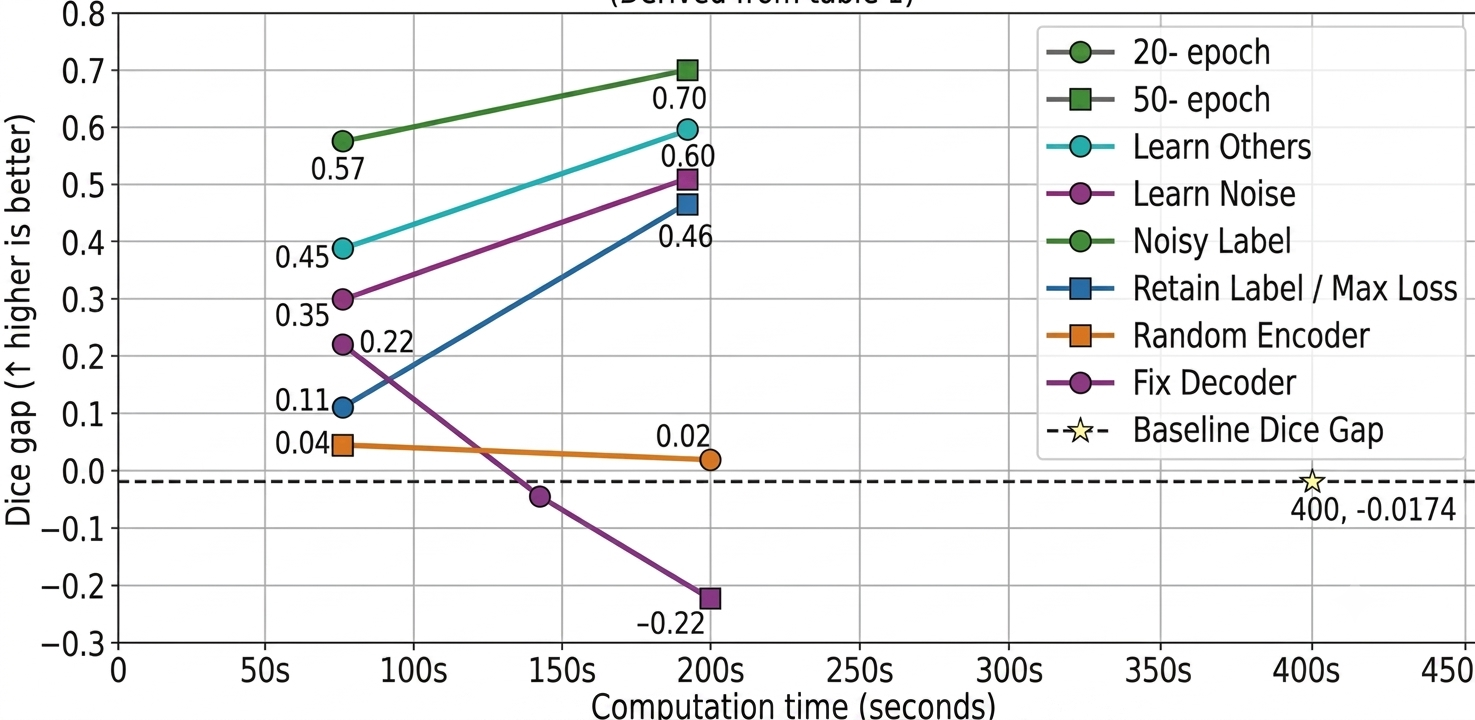}
        \caption{Performance comparison of unlearning methods in terms of dice gap and computation time}
        \label{fig:Unlearningresults}
    \end{center}
    \vspace{-0.4cm}
\end{figure}

\noindent Fig. \ref{fig:Unlearningresults} shows the dice gap values and computational time comparison across different unlearning methods for two different training horizons. Overall, the results indicate that training duration is a critical hyperparameter for direct segmentation unlearning. While several methods exhibited effective forgetting at early epochs, most degraded substantially with prolonged training. \textit{The Noisy Label strategy was the only method to achieve strong forgetting while maintaining stable retain performance, making it the most effective and robust approach among the strategies evaluated.}

\begin{table*}[t]
\centering
\caption{Results of unlearning strategies at two different epochs $20$ and $50$. $\uparrow$ indicates higher is better; $\downarrow$ indicates lower is better. Bold entries indicate the best overall configuration.}
\label{tab:unlearning}
\adjustbox{max width=\textwidth}{%
\begin{tabular}{llcccccccc}
\toprule
\textbf{Unlearning Strategies} & \textbf{Method} & \textbf{Epochs} & \textbf{Time (s)} & \textbf{Retain Dice $\uparrow$} & \textbf{Forget Dice $\downarrow$} & \textbf{Dice Gap $\uparrow$} & \textbf{Retain MAE $\downarrow$} & \textbf{Forget MAE $\uparrow$} & \textbf{MAE Gap $\uparrow$} \\

\midrule
Baseline & Fine-tune & -- & 400 & 0.9052 & 0.9226 & -0.0174 & 0.0469 & 0.0364 & -0.0104 \\

\midrule
\multirow{2}{*}{Gradient-based approaches} & \multirow{2}{*}{Retain Label / Max Loss} & 20 & 80  & 0.8600 & 0.7452 & +0.1149 & 0.0664 & 0.0885 & +0.0221 \\
                                            &                                          & 50 & 196 & 0.4559 & \textbf{0.0000} & +0.4559 & 0.2666 & \textbf{0.3075} & +0.0409 \\
\midrule
\multirow{2}{*}{Label and data manipulation methods} & \multirow{2}{*}{Noisy Label} & 20 & 81  & 0.7182 & 0.1487 & +0.5696 & 0.0970 & 0.1583 & +0.0613 \\
                                                     &                              & 50 & 197 & 0.7576 & 0.0613 & \textbf{+0.6963} & 0.0851 & 0.1761 & \textbf{+0.0910} \\
\midrule
\multirow{6}{*}{Feature-Space Interventions} & \multirow{2}{*}{Random Encoder} & 20 & 81  & \textbf{0.8527} & 0.8127 & +0.0400 & \textbf{0.0593} & 0.0658 & +0.0065 \\
                                             &                                  & 50 & 198 & 0.7908 & 0.7738 & +0.0170 & 0.0724 & 0.0750 & +0.0026 \\
                                             & \multirow{2}{*}{Learn Others}    & 20 & \textbf{77}  & 0.1286 & \textbf{0.0000} & +0.1286 & 0.1802 & 0.1909 & +0.0107 \\
                                             &                                  & 50 & \textbf{194} & 0.1230 & \textbf{0.0000} & +0.1230 & 0.1819 & 0.1911 & +0.0092 \\
                                             & \multirow{2}{*}{Learn Noise}     & 20 & 81  & 0.1750 & 0.0427 & +0.1323 & 0.1752 & 0.1854 & +0.0101 \\
                                             &                                  & 50 & 200 & 0.1260 & \textbf{0.0000} & +0.1260 & 0.1809 & 0.1901 & +0.0091 \\
\midrule
\multirow{2}{*}{Fine-tuning and selective retraining method} & \multirow{2}{*}{Fix Decoder} & 20 & 82  & 0.5933 & 0.3749 & +0.2184 & 0.1204 & 0.1416 & +0.0212 \\
                                                             &                              &50 & 198 & 0.4635 & 0.6871 & -0.2235 & 0.1625 & 0.1366 & -0.0259 \\

\bottomrule
\end{tabular}}
\vspace{-0.4cm}
\end{table*}

\section{CONCLUSION AND FUTURE WORK}

\noindent This paper investigates the performance of several approximate machine unlearning strategies using a pre-trained Med3D ResNet-50 \cite{chen2019med3d} segmentation trunk. We perform these strategies on MRBrainS18 \cite{mendrik2015mrbrains} dataset over different training epochs and provides objective quantitative results as benchmarks for volume unlearning.
The conclusion of this work shows that Noisy Label is the best strategy for this objective which achieves the largest dice gap of +0.6963 and the largest MAE gap of +0.091. All of the gradient-based methods did not provide meaningful unlearning after a short budget for learning, due to significant, catastrophic collapse in retain performance with longer budget learning \cite{jia2023model}. Moreover, the feature-space interventions and selective retraining strategies (i.e., RE, LO and FD) were too aggressive or not stable at the standard learning rate \cite{golatkar2020eternal, fan2024salun}. 

\medskip

\noindent This paper presents several new avenues for future work. In particular, formal adversarial privacy evaluations using either membership inference \cite{shokri2017membership} or model inversion attacks \cite{fredrikson2015model}  will provide rigorous evidence for the success of scrubbed residual information about forget subjects. Additionally, investigating adaptive optimization algorithms such as dynamic forget-alpha scheduling \cite{sener2018multi} and calibrated learning rate warmups in conjunction with parameter efficient fine-tuning like LoRA \cite{hu2022lora} should stabilize unlearning methods in feature space. 

\bibliographystyle{IEEEtran} 
\bibliography{reference} 

\end{document}